\documentclass{article} 
\usepackage{iclr2019_conference,times}

\usepackage{amsmath,amsfonts,bm}

\def\eqref#1{equation~\ref{#1}}

\def\1{\bm{1}}

\DeclareMathAlphabet{\mathsfit}{\encodingdefault}{\sfdefault}{m}{sl}
\SetMathAlphabet{\mathsfit}{bold}{\encodingdefault}{\sfdefault}{bx}{n}

\usepackage{hyperref}
\usepackage{url}
\usepackage{graphicx}
\usepackage{minibox}
\usepackage{makecell}
\usepackage{xcolor}
\usepackage[toc,page]{appendix}

\title{Efficient and Accurate Abnormality Mining from Radiology Reports with Customized False Positive Reduction} 

\author{Nithya Attaluri \thanks{This work was done when Nithya Attaluri was an intern at Enlitic.}\\
Cupertino High School\\
\texttt{nithya.attaluri@gmail.com} \\
\AND
Ahmed Nasir, Carolynne Powe, Harold Racz\\
Trillium Health Partners \\
\texttt{\{ahmed.nasir,carolynne.powe,harold.racz\}@thp.ca }\\
\AND
Ben Covington, Li Yao, 
Jordan Prosky, Eric Poblenz, Tobi Olatunji, Kevin Lyman\\
Enlitic Inc. \\
\texttt{\{ben,li,prosky,eric,tobi,kevin\}@enlitic.com}
}

\iclrfinalcopy
\begin{document}
\maketitle

\begin{abstract}

Obtaining datasets labeled to facilitate model development is a challenge for most machine learning tasks. The difficulty is heightened for medical imaging, where data itself is limited in accessibility and labeling requires costly time and effort by trained medical specialists. Medical imaging studies, however, are often accompanied by a medical report produced by a radiologist, identifying important features on the corresponding scan for other physicians not specifically trained in radiology. We propose a methodology for approximating image-level labels for radiology studies from associated reports using a general purpose language-processing tool for medical concept extraction and sentiment analysis, and simple manually crafted heuristics for false positive reduction. Using this approach, we label more than 175,000 Head CT studies for the presence of 33 features indicative of 11 clinically relevant conditions. For 27 of the 30 keywords that yielded positive results (3 had no occurrences), the lower bound of the confidence intervals created to estimate the percentage of accurately-labeled reports was above 85\%, with the average being above 95\%. Though noisier then manual labeling, these results suggest this method to be a viable means of labeling medical images at scale.


 

\end{abstract}

\section{Introduction}

The development of clinical applications of machine learning for use in radiology is largely dependent on the collection of sizable medical imaging datasets. While most hospitals and radiology providers maintain massive historic stores of such data, extraction of these records at scale often poses a substantial technical and logistical challenge. Furthermore, the viability of using this data for training or validation is reliant on the association of labels which are often time-intensive and costly to collect manually using specially trained radiologists or other medical professionals. 

In a typical radiology provider setting, a technician collects medical images of a patient using an imaging device such as an X-ray, CT, or MRI, and immediately stores this data in a Picture Archiving and Communication Server (PACS) in Digital Imaging and Communications in Medicine (DICOM) format. A radiologist then retrieves these images from the PACS for their interpretation and dictates a report describing their findings. Either through the use of speech to text software or manual transcription, this report is then stored as text back in the PACS or in a Radiology Information System (RIS) or Electronic Health Record (EHR). The intent of this report is to assist a referring physician or other medical professional untrained in radiology to make an informed diagnosis; as such, it ideally serves as a credible source of information regarding the presence or absence of visual patterns in the associated images.  

Given the high cost associated with manual labeling of medical images for model development, there is high utility in automating or semi-automating this process. As DICOM image data is often quite large and slow to pull from an actively-used PACS in volume, this utility is further compounded if a label can be generated in advance of image extraction for case selection and prioritization. Here, we propose the usage of a software pipeline utilizing Natural Language Processing (NLP) to approximate study-level labels for radiology images from the associated reports. This pipeline first identifies and extracts entities of clinical interest using an open source tool with a built-in mechanism that assumes the affirmation or negation of each entity via sentiment analysis. These entities are then passed through a layer of hand-crafted and clinically inspired heuristics for false positive reduction. Using this pipeline, we identify 33 terms mapped to 11 distinct clinical findings in head CT reports as a proxy for labeling the associated images.  

The methodology described herein was developed and validated alongside Trillium Health Partners (THP), a large community hospital system in Canada. As part of a study approved by THP's reseach ethics board, THP provided 176,380 radiology reports corresponding to Head CT studies interpreted at two hospital sites over 10 years. All data used for this study was reviewed and approved as de-identified by THP’s privacy team prior to analysis. 






\section{Related work}
Natural Language Processing (NLP) has been widely applied to various clinical problems with varying results \citep{wang2017clinical, karimi2017automatic,lakhani2018machine}. Over the last two decades, a number of domain-specific information extraction tools have been developed to make large volumes of clinical text more usable for downstream clinical analyis tasks. The Radiology Analysis tool (RADA) extracts medical concepts through a specialized glossary of domain concepts, attributes, and predefined grammar rules \citep{johnson1997extracting}. Breast Imaging Reporting and Data
System (BI-RADS) \citep{nassif2009information} uses a specialized lexicon, syntax analyzer, concept finder, and negation detector to extract information from Mammography reports. Medical Language Extraction and Encoding System (MEDLEE) extracted clinically significant information from reports using a controlled vocabulary and grammatical rules \citep{sevenster2012automatically}. The Lexicon Mediated Entropy Reduction (LEXIMER)
system extracts phrases with important findings through lexicon-based hierarchical decision trees \citep{kalra2004application}. Clinical Text Analysis and Knowledge Extraction System (cTAKES) from Mayo Clinic provides a dictionary-based named-entity recognizer by extracting matching concepts from the Unified Medical Language System (UMLS) Metathesaurus \citep{savova2010mayo}. Other commonly used UMLS dictionary-based tools include Health Information Text Extraction (HITEx) from Brigham and Women’s Hospital and Harvard Medical School \citep{goryachev2006suite} and MetaMap from National Library of Medicine \citep{aronson2010overview} which was the primary tool used in our work.

As noted in a 2013 Stanford paper \citep{hassanpour2016information}, although  these dictionary, vocabulary, lexicon or rule-based methods generally have high precision, they suffer from an unacceptably high proportion of false positives (low recall and low generalizability) when extracting significant findings or labels for downstream machine learning tasks from radiology reports. This may largely be due to the lexical complexity \citep{huesch2018evaluating, zech2018natural} and idiosynracies of radiology report generation between radiologists, and across institutions \citep{hassanpour2016information} and geographies as well as non-standard use of abbreviations, conjunctions, misspellings, important exclusions, differentials, suggestions, and expressions of uncertainty in radiology reports.

A number of interesting approaches have been proposed to tackle different aspects of this complex problem. Negation detection has become an important area of focus leading to the development of tools such as NegEx \citep{chapman2001simple,chapman2013extending,chapman2001evaluation} that finds `negation triggers' in clinical text and identifies the concepts or entities within the scope of the trigger. MetaMap, cTakes, and other lexicon-based tools now integrate NegEx in their pipelines \citep{aronson2010overview,savova2010mayo}. NegBio \citep{peng2018negbio} utilizes patterns on universal dependencies to identify the scope of triggers that are indicative of negation or uncertainty, compareing favorably to NegEx (an average of 9.5 percent improvement in precision and 5.1 percent in F1–score). A number of alogrithms such as DEEPEN, PyConText, and SynNeg \citep{mehrabi2015deepen, tanushi2013negation} as well as recent implementations of existing tools like cTakes and MetaMap extend or improve on NegEx  performance, also tackling different aspects of the false positive problem such as uncertainty and status detection \citep{aronson_overview_2010,savova2010mayo}. 

The lack of robust ambiguity resolution is another major weakness of traditional tools where an entity maps to several terms or concepts in the lexicon. While traditional approaches used rules and scoring systems to rank and select candidate mappings\cite{aronson2010overview}, the dominant approach to solving this problem is context-sensitive entity extraction (Word-Sense Disambiguation)\cite{finkel2004exploiting} through the use of neural embeddings and sequence models leveraging the power of deep learning or probilistic models\cite{chen2018integrating, esuli2013enhanced, gehrmann2018comparing, wu2017clinical}. These methods, when used individually or in combination with traditional approaches\cite{xia2013combining} led to cleaner concept extraction with less noisy and fewer false positives, improving the quality of labels extracted from radiology reports for downstream machine learning tasks\cite{nandhakumar2017clinically}.

\section{Materials}
\subsection{Data Access}
The data used for this study was collected from two hospital sites, hereafter denoted Site 1 and Site 2. Like most hospitals, Site 1 and Site 2 have a PACS that houses medical imaging data in DICOM (Digital Imaging and Communications in Medicine) format. DICOM is the standard file format used to transmit and store medical imaging scan data. They also have a RIS (Radiology Information System) which contains medical reports corresponding to medical images in the PACS. We were given access to 101,660 medical reports from Site 1 and 74,720 reports from Site 2.

\subsection{De-Identification of DICOM Headers and Medical Reports}





Medical regulations require that all protected health information (PHI)  is removed from medical information before the data is extracted from a hospital server and used by an external organization. This process is known as de-identification. We developed a novel anonymization process that would maximize the amount of clinically relevant information preserved for modeling and analysis.

Our anonymization algorithm was run on a server at the hospital sites. On-site PACS engineers uploaded the reports and DICOMS on the server, and we generated corresponding data that was stripped of patient information. The anonymized medical information was used in our experiments only after the hospital sites' privacy specialists cleared the data for use within our company.

Our anonymization approach leverages the tools of natural language processing (NLP) to identify PHI and replaces it with a fiducial marker indicating the category of the information that was replaced (i.e. name, age, address, medical record number, etc...). Thus, each specific instance of PHI is reduced to an abstract entity. This output format is intended to make the de-identified text amenable to information extraction algorithms.

The algorithm works in two passes. In the first pass, text corresponding to any designated PHI category is identified and parsed to extract a structured representation of the information which is then stored. In the second pass, all of the stored structured information is used to implement robust identification of PHI in the text which is then replaced with a category-specific fiducial marker.

In order to facilitate robust identification and parsing, the first pass of the algorithm ingests all available inputs including DICOM files and reports. This allows us to leverage structure in the DICOM headers where some instances of person names, ages, dates, and times are flagged as such. With unstructured text, the leverage comes from getting multiple looks at an entity such as a person's name. For example, we might catch variations that differ in terms of middle name or use of initials. In practice, this fusion of information improves the accuracy considerably.

To facilitate robust identification and replacement in the second pass, we can work with structured information accumulated across multiple first pass runs. By working with structured information, as opposed to a blacklist of terms, the algorithm can match a given instance of PHI even if it appears in a variety of different forms.

The designated PHI categories represent the types of information that the algorithm will remove. Specifically, once run through our de-identification methods, the medical reports were scrubbed of: names of persons, names of institutions, addresses, ages, dates, times, phone numbers, accession numbers, and medical record numbers. 

\section{Methods}

From the medical reports we collected, our goal was to identify those that were positive for one of 11 specific medical conditions visible on a head CT. The 11 conditions were represented in reports by a variety of specific synonyms and keywords. Hemorrhage, for instance, could be indicated by alternate spellings, such as `haemorrhage', or synonyms, including `hematoma' or `bleeding'. For the 11 conditions, we generated a list of 33 relevant keywords that we could search for in the reports (refer to Table \ref{Table 1}). Although this list of alternative words is not meant to be exhaustive, it worked sufficiently well for our purposes.

\begin{table}[ht]
\caption{General Medical Conditions and Keywords Used to Identify Medical Condition}
\label{Table 1}
\begin{center}
\renewcommand{\arraystretch}{1.3}
\begin{tabular}{|l|p{3.5in}|}
\hline
General Medical Conditions & Keywords\\
\hline
Stroke & infarct, CVA, stroke, acute ischemic event, chronic ischemic event\\
Hemorrhage & hemorrhage, haemorrhage, rupture, bleeding, hematoma\\
Encephalomalacia & encephalomalacia\\
Ischemia & ischemia, ischemic change\\
Fracture & fracture\\
Cerebral Herniation & hernia\\
Hydrocephalus & hydrocephalus\\
Tumor/Mass/Cyst & mass, malformation, arachnoid cyst, polyp, polyposis, calcification, cystic necrosis, tumor, glioblastoma, cancer, meningioma\\
Vasculopathy & aneurysm, thrombosis, thrombus, dissection, rupture\\
Neurodegenerative Disease & atrophy\\
Fluid Collection & hygroma\\
\hline
\end{tabular}
\end{center}
\end{table}

For a report to be positive for a keyword, it was required to meet two conditions:
\begin{enumerate}
\item The keyword must be present in the report
\item The keyword must not be negated by other words in the report
\end{enumerate}

\begin{figure}[ht]
\centering
\includegraphics[scale=0.30]{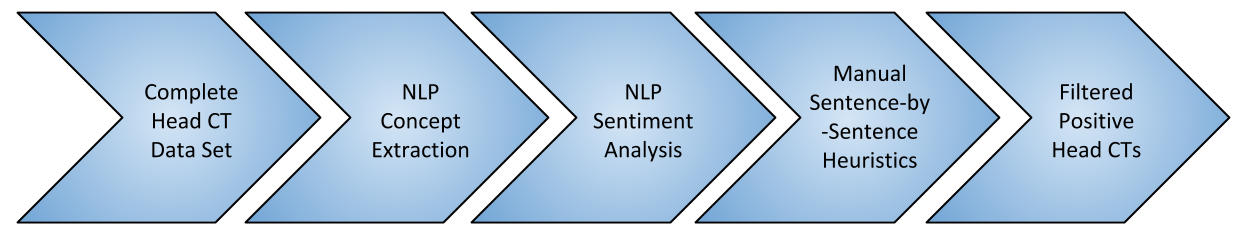} 
\caption{Layers of processing used to obtain final, auto-labeled datasets}
\label{Figure 1}
\end{figure}

We systematically evaluated every medical report based on these criteria using three primary steps (refer to Figure \ref{Figure 1}). First, a Natural Language Processor (NLP) extracted all of the medical concepts present in the text. Next, the NLP applied sentiment analysis to determine whether these conditions were positive (visible in the scan) or negative (not visible in the scan). Finally, a set of manually-generated heuristics filtered out majority of the false positives incorrectly identified by the NLP tool. 

\subsection{Processing by NLP}\label{sec_nlp}


The NLP tool we utilized for the concept extraction and sentiment analysis steps was MetaMap, a biomedical text processor developed at the National Library of Medicine (NLM) \cite{aronson_overview_2010}. MetaMap's primary function is to interpret biomedical publications and map the text to biomedical concepts defined by the Unified Medical Language System (UMLS) Metathesaurus \cite{information_metathesaurus_2009}. However, it can be adapted for processing clinical text like medical reports. 

MetaMap identifies concepts - in this case, significant medical abnormalities referred to within a report - by searching for relevant words and phrases related to that concept. By the end of the concept extraction step, all of the 33 keywords we were examining could be classified into two groups - relevant or irrelevant - for each study based on whether they were, respectively, mentioned in the report or not. 

All of the concepts identified as relevant were then subjected to sentiment analysis: every abnormality originally classified as relevant was marked as either positive or negative. Positive concepts represented medical conditions that the radiologist marked as visible in the scan. Negative concepts were those that were explicitly noted as not visible in the scan.

By the end of the sentiment analysis, the 33 keywords of interest could be classified into three groups for each scan: positive, negative, or irrelevant. For our purposes, we only considered the positive reports -  here on referred to as `NLP-Positive' - for the next step. 


\subsection{Identification of Negating `Situations'}

While spot-checking the NLP-Positive results for accuracy, we discovered that the false positive rate was far too high. Very often, the NLP marked reports as positive for certain abnormalities, when the surrounding sentence and context revealed that the abnormality was not necessarily visible or significant in the head CT. For instance, sometimes the report noted that the patient had a \textit{history} of stroke, but not that the stroke was necessarily present on the imaging scan. In other cases, the report indicated that a tumor had already been completely excised. 

The false positives identified by the NLP often fell into one of a recurring group of broad categories, which we called `negating situations'. When used in a sentence, they either indicated that the abnormality was not visible altogether or that the radiologist could not make a definite conclusion about presence of the abnormality in the scan. 


These `negating situations' are discussed in the sections below. Each is accompanied by a real, de-identified report snippet illustrating the situation.

\subsubsection{Patient History Information}

Some sentences of every medical report were allocated to describe patient history. Although these sentences showed that a patient had a significant medical condition in the past, they did not necessarily indicate that the abnormality was visible in the current scan. Other phrases, like `Clinical Indication' or `Clinical Information' were also used to demarcate the history section of the report. 

\begin{center}
\fbox{\begin{minipage}{0.8\linewidth}
\textit{History. Stroke. \\ 
No previous. \\
CT scan of the head. The CT scan is within normal limits for age. \\
$\cdots $ \\
Impression  \\
No definitive acute vascular pathology identified.
}
\end{minipage}
}
\end{center}

In the report above, the second sentence contained the only instance of stroke in the text. It is clear that this sentence is part of the history section, but the NLP marked this report as positive for `stroke' nonetheless.

\subsubsection{Intent of Scan}
Some reports declared that the intent of the scan was to `Query for' or `Rule out' an abnormality. Others put it more concisely: `CVA?' The NLP often marked these sentences as positive for an abnormality, even though they made no observation about the scan's features. \newline
\begin{center}
\fbox{\begin{minipage}{0.8\linewidth}
\textit{CLINICAL INDICATION:CVA? \newline
PREVIOUS:None \newline
FINDINGS:There is mild age-appropriate atrophy. 
There is no evidence of space occupying lesion,
acute infarction or intracranial hemorrhage.
The ventricles, subarachnoid spaces and posterior fossa structures and normal. 
There are significant changes on the left side of the face with 
destruction fractures through the orbit, sinuses and nasal bones. 
The left globe is absent. There is chronic mucosal disease in the 
remaining ethmoid sinuses and right maxillary sinus and in the ethmoid 
air cells on the left side. The frontal sinuses appear completely opacified. \newline
INTERPRETATION:No acute brain parenchymal abnormality is identified. 
There is chronic sinusitis as well as evidence of previous injury or 
destruction of the orbit and sinuses and nasal bones on the left side.}
\end{minipage}
}
\end{center}

This sample report was marked NLP-Positive for CVA despite the only occurrence of CVA in the report text being in the first sentence describing the intention of performing the head CT. The only abnormalities indicated by the report are fractures, trauma, sinitus, and atrophy, none of which are directly related to CVA or stroke-related conditions.

\subsubsection{Inconclusive Diagnoses}
Radiologists sometimes mentioned that it was `difficult to say/determine/comment' on the presence of an abnormality. These sentences could not be used to definitively mark a scan as positive.

\begin{center}
\fbox{\begin{minipage}{0.8\linewidth}
\textit{History: Rule out bleed. \newline 
Findings: There is mild motion artifact on the examination degrades image quality. $ \cdots $ There is loss of gray-white matter differentiation within the right cerebral hemisphere. It is difficult to determine if this is simply due to motion artifact or could represent an area of recent infarction. There is no intracerebral hemorrhage. There is mild diffuse age related atrophy. The ventricles are otherwise unremarkable. The brainstem and posterior fossa appear unremarkable. No intracranial mass effect is seen. \newline
Impression: Limited study due to motion artifact. An area of recent infarction within the right cerebral hemisphere is difficult to exclude. Clinical correlation is suggested. If required, a repeat CT scan may be helpful. There is no evidence of recent intracranial hemorrhage.}
\end{minipage}
}
\end{center}

The report above was marked NLP-Positive for `recent infarct' despite the radiologist's note that the feature on the scan potentially representing an infarct could also be the result of a motion artifact. This statement is not sufficient to mark the report as positive for infarct. 

\subsubsection{Patient Family Information}
There were a variety of ways in which a report indicated that a patient had a family history of a condition: `Mother had hemorrhage', `Sister had cancer`. These statements, although marked as NLP-positive, could not tell us anything concrete about the features visible in the patient's medical imaging scan.

\begin{center}
\fbox{\begin{minipage}{0.8\linewidth}
\textit{CT HEAD WITHOUT CONTRAST \newline 
Indication: Recurrent severe headache. Family history of brain tumor. \newline 
Technique: A nonenhanced CT scan of the brain was performed. \newline 
Comparison: \{\{DATETIME\}\} \newline 
Findings: There is no evidence of an intracranial hemorrhage, infarct, mass or mass effect. There is mild generalized atrophy. There is normal gray-white differentiation. Structures are midline. The craniocervical junction is normal. The posterior fossa is unremarkable.The cranial vault, and mastoids are unremarkable. Mucosal thickening is noted in both maxillary sinuses. \newline 
IMPRESSION:No acute intracranial abnormalities. Bilateral maxillary sinus mucosal thickening }
\end{minipage}
}
\end{center}

The above report was marked NLP-Positive for `tumour of the brain' although the only mention of a tumor is to note that there is a family history of brain tumor. There is nothing indicating the presence of a tumor in the rest of the report. In fact, the radiologist mentions that there is no evidence of mass or mass effect visible on the scan. 

\subsubsection{Low Sensitivity/Resolution of Scan}
Sometimes, radiologists noted in the reports that the CTs performed were not sensitive enough to detect the abnormality of interest. For instance, discrete vascular abnormalities were sometimes not visualized as clearly as desired on a regular CT scan, and often doctors recommended a CT angiogram to make a better diagnosis. It was inaccurate to conclude that reports with such referrals were positive for an abnormality. \newline

\begin{center}
\fbox{\begin{minipage}{0.8\linewidth}
\textit{HISTORY: Parkinsonism. Rule out vascular disease. \newline
COMPARISON: None \newline
FINDINGS: Mild generalized atrophy. No significant microangiopathic changes noted. No acute infarction, hemorrhage or masses. There is minor calcification of the intracerebral ICA at the bilaterally. Ventricles and sulci are age-appropriate. Orbits and soft tissues unremarkable. Paranasal sinuses are clear. No calvarial lesion. \newline
INTERPRETATION: Generalized atrophy. No other significant abnormalities. If there is a concern for a vascular malformation, CT angiogram or MRI is recommended.}
\end{minipage}
}
\end{center}

This sample report was marked NLP-Positive for vascular malformation, even though the only mention of the abnormality is when the radiologist suggests a CT angiogram or MRI to check for a vascular malformation. This sentence does not conclusively indicate the presence of malformation. 

\subsubsection{Tumor Specific Negators}
Specifically for tumors, sentences suggesting tumor cavities or residues, or procedures indicating varying degrees of resection, craniotomy, and debulking usually indicated that the tumor was no longer prominently visible on the CT scan if not removed completely. The NLP incorrectly coded such scans as positive for tumor. \newline

\begin{center}
\fbox{\begin{minipage}{0.8\linewidth}
\textit{History: Craniotomy for tumor \newline
Previous: \{\{DATETIME\}\}. \newline
Findings: Left frontal craniotomy. Small amount of postoperative air and hemorrhage with adjacent edema in the left frontal lobe at the site of surgery. No significant mass effect. No other significant abnormality.}
\end{minipage}
}
\end{center}

The report above was incorrectly marked NLP-Positive for tumor despite a note by the radiologist that the patient's craniotomy  left no significant mass effect. 

\subsubsection{Aneurysm Specific Negators}
Although noting the presence of aneurysm `clippings', `coiling', or post-operative metal `artifacts' in reports, indicated that the patient had an aneurysm at one point, scans with these features showed that the aneurysm had been treated, and thus the NLP's label of positive for a clinically significant aneurysm was incorrect for these scans. \newline

\begin{center}
\fbox{\begin{minipage}{0.8\linewidth}
\textit{CT HEAD UNENHANCED: \newline
CLINICAL HISTORY: Vertigo for 3 weeks. No previous examinations are available for comparison. There is evidence of a frontal craniotomy with a dense metallic artifact identified in the midline, suggestive of anterior cerebral artery aneurysm clipping. This is associated with right frontal encephalomalacia. It is chronic. No evidence of other focal disease is identified with no evidence of brain stem abnormalities. Soft tissue windows are unremarkable. \newline
OPINION: Post surgical changes. No evidence of acute intracranial processes.}
\end{minipage}
}
\end{center}

This sample report was marked NLP-Positive for `anterior cerebral artery aneurysm'. Although the patient once had this type of aneurysm, the presence of a clip on the aneurysm indicates that it has already been treated and/or cannot be seen on the scan due to local scatter artifact. 

\subsubsection{Treatment Already Administered} 
In some situations, especially stroke, the medical imaging scan was performed after the patient had received some treatment for their condition. Because post-treatment states for stroke can vary considerably, we could not conclude anything about the severity of the patient's stroke unless the radiologist mentioned specific details later on in the report. \newline

\begin{center}
\fbox{\begin{minipage}{0.8\linewidth}
\textit{This patient is post TPA treatment for acute stroke. This study is essentially within normal limits. There is no evidence of any hemorrhage or acute stroke. There is no subdural. There is no hydrocephalus.}
\end{minipage}
}
\end{center}

The report above was marked NLP-Positive for `acute stroke'. However, the report states that the study is nearly normal and that there is no evidence of acute stroke on the scan. 

\subsection{False positive reduction}\label{sec_fpr}
Clearly, our NLP was not sufficiently accurate for our purposes. Thus, in our second step, we generated heuristics that would examine a report sentence by sentence. Sentences were delimited by a period followed by either a blank space or a new line character \verb|\n|. When it encountered a sentence that seemed to fall into one of these `negating situations', our program overlooked the sentence instead of marking the report positive for that abnormality. 

It is important to clarify that, in this step, a report was never marked negative for an abnormality. We iterated through each sentence of a report, searching for a sentence that indicated that an abnormality was present. If we could find such a sentence, the report was marked positive for that abnormality. If we could find multiple sentences indicating that multiple abnormalities were present, we marked the report positive for each abnormality. If there was no such sentence in the entire report, the report was not marked in any way. 

This logic meant that a report with a negative or ambiguous sentence such as “Query for hemorrhage.” followed by “Hemorrhage is present.” later in the report was still marked positive for hemorrhage. However, if “Query for hemorrhage.” was the only sentence containing “hemorrhage” in the report, the report would be marked negative for the condition.

For a sentence to be a positive indicator for an abnormality, it had to meet two criteria. First, it had to contain the name of the abnormality of interest. And second, it could not contain any one of a list of `excluded words'. These words (listed in Table \ref{Table 2}) were generated based on manual examination of the `negating situations' described above. 

These conditions were perhaps more stringent than necessary, but we sacrificed the potentially larger set of labeled data for more accurately-labeled results. 

\begin{table}[ht]
\caption{List of excluded words and situations prompting their exclusion. Each of these excluded words is derived from one of the `negating situations' described earlier. The first eight categories of words were applied when searching for all labels; the remaining eight were applied for only the labels to which they were relevant. The specific excluded words not discussed above are explained here: `Staging' is a cancer-specific word to indicate intent of scan; used when patient already has cancer and radiologist is checking for metastasis. `MRI' and `CT angiogram' are often found in statements referring the patient for another test, usually if the current scan is not enough to make a diagnosis. `Treatment' and `insensitive' are usually found in sentences describing treatment already administered for abnormality.}
\label{Table 2}
\begin{center}
\begin{tabular}{|c|c|}
\hline
Situations & Corresponding ``Excluded'' Words\\
\hline
\hline
Universal Negators & \makecell{no, neither, ruled out, unlikely, \\without, less likely, absent, \\lack, don't, can't, cannot, exclude, never, \\ less likely, denies, denied, resolved, negative}\\
\hline
Inconclusive Diagnoses & \makecell{uncertain, difficult to determine, difficult to say, \\difficult to comment, should be excluded, \\could be excluded, must be excluded, inconsistent}\\
\hline
Patient History Information & prior, history, indication, previous, earlier\\
Intent of Scan & query, ?, rule out\\
\hline
Low Scan Resolution/Sensitivity & resolution, sensitivity, insensitive\\
\hline
Patient Family & brother, sister, mother, father, family\\
\hline
Resolved/Non-Severe Abnormalities & old, age related, age-related, repair, removed, decrease\\
\hline
Tumor Specific Words & \makecell{postop, craniotomy, resection, cavity, residual, \\
pseudotumor, debulked}\\
\hline
Aneurysm Specific Words & clip, metal, artifact, coil\\
\hline
Glioblastoma Specific Words & postop, craniotomy, resection\\
\hline
Cancer Specific Words & staging   \\
\hline
CVA Specific Words & clinical information  \\
\hline
Stroke Specific Words & \makecell{clinical information, previous, \\MRI, CT angiogram, treatment}\\
\hline
Malformation Specific Words & MRI, CT angiogram\\
\hline
Specific Words for Any Acute Condition & subacute, sub-acute\\
\hline
\end{tabular}
\end{center}
\end{table}

\section{Results}
By the end of our analysis, we had compiled a list of reports that corresponded to head CT's indicating one or more significant abnormalities. Ultimately, our methodology identified 39,028 reports as positive for at least one abnormality from Site 1, and 25,704 reports as positive from Site 2. Some reports were positive for multiple abnormalities. 

\subsection{Spot Checking \& Confidence Intervals}

We utilized statistical sampling techniques to determine the accuracy levels of the lists of reports we collected. For all of the reports positive for one specific abnormality - for instance, hemorrhage - we randomly sampled around thirty to fifty reports and presented them to human radiologists to arbitrate whether the model-generated abnormality was indeed present. This process was repeated for each of the 33 abnormalities of interest.

Based on the percentage of accurately identified reports in a sample, we were able to generate confidence intervals to obtain an interval estimate of the true percentage of accurately identified reports in the entire population of positive reports for a single abnormality. 

To calculate confidence intervals, we computed the percentage of correct reports by the model, calling this $ \hat{p} $. We then computed the Standard Error of $ \hat{p} $ using Equ. (1), where $ n $ is the total number of reports sampled:
\begin{equation} 
SE(\hat{p}) = \sqrt{\hat{p}(1 - \hat{p})} \cdot f(n)
\end{equation}

where $ f(n) $ is called the finite population correction factor and can be used to produce a more accurate confidence interval when the total population size ($ N $) is known and the sample size ($ n $) is relatively large: 
\begin{equation}
f(n) = \sqrt{\frac{N - n}{N - 1}}
\end{equation}

We used t-based confidence intervals due to the population standard deviations being unknown. For a confidence level of 95\%, the corresponding t-score value was 2.04 since the majority of our sample sizes were around 30. A confidence interval for the population percentage of correct reports is of the form:
\begin{equation}
\hat{p} \pm t_{n,q} \cdot \frac{SE(\hat{p})}{\sqrt{n}}
\end{equation}

Three keywords, \verb|arachnoid cyst|, \verb|haemorrhage| (alternate spelling of `hemorrhage'), and \verb|chronic ischemic event|, did not yield any positive results. However, once confidence intervals were calculated for the remaining abnormalities, we found that at least 85\% or more of the positive reports were accurately labeled, with the exception of the lists for \verb|cva|, \verb|stroke|, and \verb|acute ischemic event|. These three conditions, however, are often broad diagnoses that often manifest into more specific features seen on a CT scan. These keywords may not be the best descriptors of physical features on a head CT. 


\subsection{With and without false positive reduction}
To differentiate the results with and without the proposed false positive reduction, we tag medical conditions detected by standard NLP tools (in Section \ref{sec_nlp}) as `NLP-positive'. By contrast, we name those further processed by our proposed false positive reduction (in Section \ref{sec_fpr}) as `Final-positive'.

In Table \ref{Table 3}, we show results in both precision estimate and its confidence interval on conditions of both NLP-positive and Final-positive. It is immediately obvious that additional false positive reduction improves significantly the performance of positive condition identification. Table \ref{Table 4} and Table \ref{Table 5} in the Appendix include the full set of conditions of interests from both data sites.

\begin{table}[th]
\small
\caption{Accuracy calculations for select abnormalities after NLP-processing and false positive reduction. The first set of three columns represents number of reports marked NLP-Positive for a certain abnormality, number of accurately labeled reports in a set of reports sampled from these NLP-Positive reports, and corresponding confidence interval (at $ \alpha = 95\% $). Second set of three columns represents same set of calculations for reports marked positive at the end of the additional processing with heuristics. The data in the top half of the table represent Site 1 calculations, while data in the bottom half is from Site 2. Precision is the number of accurately-labeled reports divided by the number of reports sampled}
\label{Table 3}
\begin{center}
\begin{tabular}{ | c || c | c | c || c | c | c |}
\hline
\small Keyword & \small NLP-pos. & \small precision  &  \small C.I. & \small Final-pos. & \small precision & \small C.I.\\
\hline
atrophy & 36296 & 25/31 & [0.662, 0.951] & 19052 & 51/52 & [0.942, 1.000]\\
calcification & 5967 & 26/31 & [0.704, 0.973] & 2935 & 33/33 & [1.000, 1.000]\\
encephalomalacia & 2296 & 27/31 & [0.749, 0.993] & 959 & 32/32 & [1.000, 1.000]\\
hemorrhage & 9709 & 24/30 & [0.651, 0.949] & 3678 & 31/33 & [0.855, 1.000]\\
infarct & 25205 & 25/30 & [0.695, 0.972] & 9976 & 33/33 & [1.000, 1.000]\\
ischemia & 3991 & 33/41 & [0.679, 0.931] & 2866 & 34/34 & [1.000, 1.000]\\
mass & 9548 & 20/30 & [0.491, 0.842] & 3959 & 33/33 & [1.000, 1.000]\\
rupture & 139 & 23/31 & [0.600, 0.884] & 61 & 33/35 & [0.890, 0.996]\\
\hline
\hline
atrophy & 54679 & 23/30 & [0.609, 0.924] & 11421 & 30/30 & [1.000, 1.000]\\
calcification & 5967 & 27/30 & [0.789, 1.000] & 2800 & 30/30 & [1.000, 1.000]\\
encephalomalacia & 4379 & 27/30 & [0.789, 1.000] & 889 & 30/30 & [1.000, 1.000]\\
hemorrhage & 13460 & 26/30 & [0.740, 0.993] & 1396 & 48/51 & [0.875, 1.000]\\
infarct & 25205 & 23/30 & [0.609, 0.924] & 6531 & 47/48 & [0.937, 1.000]\\
ischemia & 6463 & 26/30 & [0.740, 0.993] & 1891 & 30/31 & [0.904, 1.000]\\
mass & 9548 & 24/30 & [0.651, 0.949] & 2648 & 44/47 & [0.864, 1.000]\\
rupture & 192 & 22/30 & [0.582, 0.885] & 24 & 24/24 & 1.000\\
\hline
\end{tabular}
\end{center} 
\end{table}

\section{Discussion} 


Our results indicate that it is, in fact, possible to use medical reports to automatically generate a labeled data set of medical imaging scans with high accuracy. As shown in Table \ref{Table 3}, the accuracy of the report labels increases after the false positive reduction step, indicating that the generated heuristics can compensate for lacking NLP performance. 

Reporting styles tend to differ across hospital sites. Our heuristics for false positive reduction were tailored specifically for Site 1, but the similar accuracy values and confidence intervals across the two sites indicate that the heuristics and processing applied to the reports are robust and generalizable. We anticipate that our collection of `excluded words' will be helpful for a similar study on a different set of medical reports. However, we believe that the general, `negating situations' we identified, which caused most of the false positives coded by the NLP, will be more valuable. These will aid other research groups in developing their own unique sets of `excluded words' that are customized to the writing styles of different medical data sets. 

By decreasing the need for manual data-labeling ventures, the image-level labels we were able to automatically generate will allow for labeling on a scale that has not been possible thus far. The labels allow for quick identification of studies of interest, so that experimenters can pull only relevant studies from hospital PACS systems. 

However, this methodology also has its limitations. It is necessary to ensure that the reports generated by the hospital are all-encompassing and accurate. It is possible that a radiologist may miss an abnormality or that a scan has been performed with a specific motivation for which the presence of a certain abnormality is irrelevant. In such cases our assumption that radiology reports are gold-standard datasets does not hold. 

In addition, obtaining access to an individual hospital's records is a difficult process. If the NLP tool used has a false negative rate that is too high or the heuristics employed are far too strict, the number of significant reports identified may be much lower than the true value. Without maximizing the amount of significant data pulled from the hospital servers, experimenters may be forced to go through the negotiation process to obtain hospital data more frequently than necessary. 

Our current methodology does not work towards minimizing the false negative rate. Thus, while our true positive rates are sufficiently high, we overlook a great deal of significant data which is wasteful when data is so scarce. In the future, we hope to maximize the number of significant scans we identify without sacrificing the high levels of accuracy that we currently show. We also understand that there is great value to identifying imaging scans that appear normal. This is an additional avenue for expansion of our auto-labeling methodology. 






\bibliography{main}

\begin{thebibliography}{28}
\providecommand{\natexlab}[1]{#1}
\providecommand{\url}[1]{\texttt{#1}}
\expandafter\ifx\csname urlstyle\endcsname\relax
  \providecommand{\doi}[1]{doi: #1}\else
  \providecommand{\doi}{doi: \begingroup \urlstyle{rm}\Url}\fi

\bibitem[Aronson \& Lang(2010{\natexlab{a}})Aronson and
  Lang]{aronson2010overview}
Alan~R Aronson and Fran{\c{c}}ois-Michel Lang.
\newblock An overview of metamap: historical perspective and recent advances.
\newblock \emph{Journal of the American Medical Informatics Association},
  17\penalty0 (3):\penalty0 229--236, 2010{\natexlab{a}}.

\bibitem[Aronson \& Lang(2010{\natexlab{b}})Aronson and
  Lang]{aronson_overview_2010}
Alan~R Aronson and François-Michel Lang.
\newblock An overview of {MetaMap}: historical perspective and recent advances.
\newblock \emph{Journal of the American Medical Informatics Association},
  17\penalty0 (3):\penalty0 229--236, May 2010{\natexlab{b}}.
\newblock ISSN 1067-5027, 1527-974X.
\newblock \doi{10.1136/jamia.2009.002733}.
\newblock URL
  \url{https://academic.oup.com/jamia/article-lookup/doi/10.1136/jamia.2009.002733}.

\bibitem[Chapman et~al.(2001{\natexlab{a}})Chapman, Bridewell, Hanbury, Cooper,
  and Buchanan]{chapman2001evaluation}
Wendy~W Chapman, Will Bridewell, Paul Hanbury, Gregory~F Cooper, and Bruce~G
  Buchanan.
\newblock Evaluation of negation phrases in narrative clinical reports.
\newblock In \emph{Proceedings of the AMIA Symposium}, pp.\  105. American
  Medical Informatics Association, 2001{\natexlab{a}}.

\bibitem[Chapman et~al.(2001{\natexlab{b}})Chapman, Bridewell, Hanbury, Cooper,
  and Buchanan]{chapman2001simple}
Wendy~W Chapman, Will Bridewell, Paul Hanbury, Gregory~F Cooper, and Bruce~G
  Buchanan.
\newblock A simple algorithm for identifying negated findings and diseases in
  discharge summaries.
\newblock \emph{Journal of biomedical informatics}, 34\penalty0 (5):\penalty0
  301--310, 2001{\natexlab{b}}.

\bibitem[Chapman et~al.(2013)Chapman, Hilert, Velupillai, Kvist, Skeppstedt,
  Chapman, Conway, Tharp, Mowery, and Deleger]{chapman2013extending}
Wendy~W Chapman, Dieter Hilert, Sumithra Velupillai, Maria Kvist, Maria
  Skeppstedt, Brian~E Chapman, Michael Conway, Melissa Tharp, Danielle~L
  Mowery, and Louise Deleger.
\newblock Extending the negex lexicon for multiple languages.
\newblock \emph{Studies in health technology and informatics}, 192:\penalty0
  677, 2013.

\bibitem[Chen et~al.(2018)Chen, Zafar, Galperin-Aizenberg, and
  Cook]{chen2018integrating}
Po-Hao Chen, Hanna Zafar, Maya Galperin-Aizenberg, and Tessa Cook.
\newblock Integrating natural language processing and machine learning
  algorithms to categorize oncologic response in radiology reports.
\newblock \emph{Journal of digital imaging}, 31\penalty0 (2):\penalty0
  178--184, 2018.

\bibitem[Esuli et~al.(2013)Esuli, Marcheggiani, and
  Sebastiani]{esuli2013enhanced}
Andrea Esuli, Diego Marcheggiani, and Fabrizio Sebastiani.
\newblock An enhanced crfs-based system for information extraction from
  radiology reports.
\newblock \emph{Journal of biomedical informatics}, 46\penalty0 (3):\penalty0
  425--435, 2013.

\bibitem[Finkel et~al.(2004)Finkel, Dingare, Nguyen, Nissim, Manning, and
  Sinclair]{finkel2004exploiting}
Jenny Finkel, Shipra Dingare, Huy Nguyen, Malvina Nissim, Christopher Manning,
  and Gail Sinclair.
\newblock Exploiting context for biomedical entity recognition: From syntax to
  the web.
\newblock In \emph{Proceedings of the International Joint Workshop on Natural
  Language Processing in Biomedicine and its Applications}, pp.\  88--91.
  Association for Computational Linguistics, 2004.

\bibitem[Gehrmann et~al.(2018)Gehrmann, Dernoncourt, Li, Carlson, Wu, Welt,
  Foote~Jr, Moseley, Grant, Tyler, et~al.]{gehrmann2018comparing}
Sebastian Gehrmann, Franck Dernoncourt, Yeran Li, Eric~T Carlson, Joy~T Wu,
  Jonathan Welt, John Foote~Jr, Edward~T Moseley, David~W Grant, Patrick~D
  Tyler, et~al.
\newblock Comparing deep learning and concept extraction based methods for
  patient phenotyping from clinical narratives.
\newblock \emph{PloS one}, 13\penalty0 (2):\penalty0 e0192360, 2018.

\bibitem[Goryachev et~al.(2006)Goryachev, Sordo, and Zeng]{goryachev2006suite}
Sergey Goryachev, Margarita Sordo, and Qing~T Zeng.
\newblock A suite of natural language processing tools developed for the i2b2
  project.
\newblock In \emph{AMIA Annual Symposium Proceedings}, volume 2006, pp.\  931.
  American Medical Informatics Association, 2006.

\bibitem[Hassanpour \& Langlotz(2016)Hassanpour and
  Langlotz]{hassanpour2016information}
Saeed Hassanpour and Curtis~P Langlotz.
\newblock Information extraction from multi-institutional radiology reports.
\newblock \emph{Artificial intelligence in medicine}, 66:\penalty0 29--39,
  2016.

\bibitem[Huesch et~al.(2018)Huesch, Cherian, Labib, and
  Mahraj]{huesch2018evaluating}
Marco~D Huesch, Rekha Cherian, Sam Labib, and Rickhesvar Mahraj.
\newblock Evaluating report text variation and informativeness: natural
  language processing of ct chest imaging for pulmonary embolism.
\newblock \emph{Journal of the American College of Radiology}, 15\penalty0
  (3):\penalty0 554--562, 2018.

\bibitem[Information et~al.(2009)Information, Pike, MD, and
  Usa]{information_metathesaurus_2009}
National Center for~Biotechnology Information, U.~S. National Library of
  Medicine 8600~Rockville Pike, Bethesda MD, and 20894 Usa.
\newblock \emph{Metathesaurus}.
\newblock National Library of Medicine (US), September 2009.
\newblock URL \url{https://www.ncbi.nlm.nih.gov/books/NBK9684/}.

\bibitem[Johnson et~al.(1997)Johnson, Taira, Cardenas, and
  Aberle]{johnson1997extracting}
David~B Johnson, Ricky~K Taira, Alfonso~F Cardenas, and Denise~R Aberle.
\newblock Extracting information from free text radiology reports.
\newblock \emph{International Journal on Digital Libraries}, 1\penalty0
  (3):\penalty0 297--308, 1997.

\bibitem[Kalra et~al.(2004)Kalra, Dreyer, Maher, et~al.]{kalra2004application}
MK~Kalra, KJ~Dreyer, MM~Maher, et~al.
\newblock Application of information theory-based, search engine, leximer for
  automatic classification of unstructured radiology reports.
\newblock \emph{Chicago: infoRAD exhibit, RSNA}, 2004.

\bibitem[Karimi et~al.(2017)Karimi, Dai, Hassanzadeh, and
  Nguyen]{karimi2017automatic}
Sarvnaz Karimi, Xiang Dai, Hamedh Hassanzadeh, and Anthony Nguyen.
\newblock Automatic diagnosis coding of radiology reports: a comparison of deep
  learning and conventional classification methods.
\newblock \emph{BioNLP 2017}, pp.\  328--332, 2017.

\bibitem[Lakhani et~al.(2018)Lakhani, Prater, Hutson, Andriole, Dreyer, Morey,
  Prevedello, Clark, Geis, Itri, et~al.]{lakhani2018machine}
Paras Lakhani, Adam~B Prater, R~Kent Hutson, Kathy~P Andriole, Keith~J Dreyer,
  Jose Morey, Luciano~M Prevedello, Toshi~J Clark, J~Raymond Geis, Jason~N
  Itri, et~al.
\newblock Machine learning in radiology: applications beyond image
  interpretation.
\newblock \emph{Journal of the American College of Radiology}, 15\penalty0
  (2):\penalty0 350--359, 2018.

\bibitem[Mehrabi et~al.(2015)Mehrabi, Krishnan, Sohn, Roch, Schmidt, Kesterson,
  Beesley, Dexter, Schmidt, Liu, et~al.]{mehrabi2015deepen}
Saeed Mehrabi, Anand Krishnan, Sunghwan Sohn, Alexandra~M Roch, Heidi Schmidt,
  Joe Kesterson, Chris Beesley, Paul Dexter, C~Max Schmidt, Hongfang Liu,
  et~al.
\newblock Deepen: A negation detection system for clinical text incorporating
  dependency relation into negex.
\newblock \emph{Journal of biomedical informatics}, 54:\penalty0 213--219,
  2015.

\bibitem[Nandhakumar et~al.(2017)Nandhakumar, Sherkat, Milios, Gu, and
  Butler]{nandhakumar2017clinically}
Nidhin Nandhakumar, Ehsan Sherkat, Evangelos~E Milios, Hong Gu, and Michael
  Butler.
\newblock Clinically significant information extraction from radiology reports.
\newblock In \emph{Proceedings of the 2017 ACM Symposium on Document
  Engineering}, pp.\  153--162. ACM, 2017.

\bibitem[Nassif et~al.(2009)Nassif, Woods, Burnside, Ayvaci, Shavlik, and
  Page]{nassif2009information}
Houssam Nassif, Ryan Woods, Elizabeth Burnside, Mehmet Ayvaci, Jude Shavlik,
  and David Page.
\newblock Information extraction for clinical data mining: a mammography case
  study.
\newblock In \emph{Data Mining Workshops, 2009. ICDMW'09. IEEE International
  Conference on}, pp.\  37--42. IEEE, 2009.

\bibitem[Peng et~al.(2018)Peng, Wang, Lu, Bagheri, Summers, and
  Lu]{peng2018negbio}
Yifan Peng, Xiaosong Wang, Le~Lu, Mohammadhadi Bagheri, Ronald Summers, and
  Zhiyong Lu.
\newblock Negbio: a high-performance tool for negation and uncertainty
  detection in radiology reports.
\newblock \emph{AMIA Summits on Translational Science Proceedings},
  2017:\penalty0 188, 2018.

\bibitem[Savova et~al.(2010)Savova, Masanz, Ogren, Zheng, Sohn, Kipper-Schuler,
  and Chute]{savova2010mayo}
Guergana~K Savova, James~J Masanz, Philip~V Ogren, Jiaping Zheng, Sunghwan
  Sohn, Karin~C Kipper-Schuler, and Christopher~G Chute.
\newblock Mayo clinical text analysis and knowledge extraction system (ctakes):
  architecture, component evaluation and applications.
\newblock \emph{Journal of the American Medical Informatics Association},
  17\penalty0 (5):\penalty0 507--513, 2010.

\bibitem[Sevenster et~al.(2012)Sevenster, Van~Ommering, and
  Qian]{sevenster2012automatically}
Merlijn Sevenster, Rob Van~Ommering, and Yuechen Qian.
\newblock Automatically correlating clinical findings and body locations in
  radiology reports using medlee.
\newblock \emph{Journal of digital imaging}, 25\penalty0 (2):\penalty0
  240--249, 2012.

\bibitem[Tanushi et~al.(2013)Tanushi, Dalianis, Duneld, Kvist, Skeppstedt, and
  Velupillai]{tanushi2013negation}
Hideyuki Tanushi, Hercules Dalianis, Martin Duneld, Maria Kvist, Maria
  Skeppstedt, and Sumithra Velupillai.
\newblock Negation scope delimitation in clinical text using three approaches:
  Negex, pycontextnlp and synneg.
\newblock In \emph{19th Nordic Conference of Computational Linguistics
  (NODALIDA 2013), May 22-24, 2013, Oslo, Norway}, pp.\  387--474.
  Link{\"o}ping University Electronic Press, 2013.

\bibitem[Wang et~al.(2017)Wang, Wang, Rastegar-Mojarad, Moon, Shen, Afzal, Liu,
  Zeng, Mehrabi, Sohn, et~al.]{wang2017clinical}
Yanshan Wang, Liwei Wang, Majid Rastegar-Mojarad, Sungrim Moon, Feichen Shen,
  Naveed Afzal, Sijia Liu, Yuqun Zeng, Saeed Mehrabi, Sunghwan Sohn, et~al.
\newblock Clinical information extraction applications: a literature review.
\newblock \emph{Journal of biomedical informatics}, 2017.

\bibitem[Wu et~al.(2017)Wu, Jiang, Xu, Zhi, and Xu]{wu2017clinical}
Yonghui Wu, Min Jiang, Jun Xu, Degui Zhi, and Hua Xu.
\newblock Clinical named entity recognition using deep learning models.
\newblock In \emph{AMIA Annual Symposium Proceedings}, volume 2017, pp.\  1812.
  American Medical Informatics Association, 2017.

\bibitem[Xia et~al.(2013)Xia, Zhong, Liu, Tan, Na, Hu, and
  Huang]{xia2013combining}
Yunqing Xia, Xiaoshi Zhong, Peng Liu, Cheng Tan, Sen Na, Qinan Hu, and Yaohai
  Huang.
\newblock Combining metamap and ctakes in disorder recognition: Thcib at clef
  ehealth lab 2013 task 1.
\newblock In \emph{CLEF (Working Notes)}, 2013.

\bibitem[Zech et~al.(2018)Zech, Pain, Titano, Badgeley, Schefflein, Su, Costa,
  Bederson, Lehar, and Oermann]{zech2018natural}
John Zech, Margaret Pain, Joseph Titano, Marcus Badgeley, Javin Schefflein,
  Andres Su, Anthony Costa, Joshua Bederson, Joseph Lehar, and Eric~Karl
  Oermann.
\newblock Natural language--based machine learning models for the annotation of
  clinical radiology reports.
\newblock \emph{Radiology}, 287\penalty0 (2):\penalty0 570--580, 2018.

\end{thebibliography}
\bibliographystyle{iclr2019_conference}

\clearpage
\appendix
\section{Appendix}
\begin{table}[ht]
\caption{Samples and confidence intervals generated per abnormality after false positive reduction for Site 1}
\label{Table 4}
\begin{center}
\renewcommand{\arraystretch}{1.3}
\begin{tabular}{ | c| c | c | c|}
\hline
Keyword & \# Pos. Reports & precision & C.I. ($ \alpha = 95\% $)\\
\hline
acute ischemic event \footnote[1] & 2 & 1/2 & 0.500\\
aneurysm & 552 & 30/30 & [1.000, 1.000]\\
arachnoid cyst \footnote[2] & 0 & N/A & N/A\\
atrophy & 19052 & 51/52 & [0.942, 1.000]\\
bleeding & 66 & 30/30 & [1.000, 1.000]\\
calcification & 2935 & 33/33 & [1.000, 1.000]\\
cancer & 3 & 3/3 & 1.000\\
chronic ischemic event & 1 & 1/1 & 1.000\\
cva & 8 & 7/8 & 0.875\\
cystic necrosis & 1 & 1/1 & 1.000\\
dissection & 80 & 32/33 & [0.923, 1.000]\\
encephalomalacia & 959 & 32/32 & [1.000, 1.000]\\
fracture & 903 & 33/34 & [0.913, 1.000]\\
glioblastoma & 19 & 18/19 & 0.947\\
haemorrhage \footnote[2] & 0 & N/A & N/A\\
hematoma & 2927 & 35/36 & [0.917, 1.000]\\
hemorrhage & 3678 & 31/33 & [0.855, 1.000]\\
hernia & 62 & 33/33 & [1.000, 1.000]\\
hydrocephalus & 610 & 33/34 & [0.913, 1.000]\\
hygroma & 275 & 36/36 & [1.000, 1.000]\\
infarct & 9976 & 33/33 & [1.000, 1.000]\\
ischemia & 2866 & 34/34 & [1.000, 1.000]\\
ischemic change & 2981 & 33/33 & [1.000, 1.000]\\
malformation & 38 & 31/32 & [0.943, 0.994]\\
mass & 3959 & 33/33 & [1.000, 1.000]\\
meningioma & 1036 & 33/34 & [0.912, 1.000]\\
polyp & 1827 & 34/34 & [1.000, 1.000]\\
polyposis & 547 & 32/32 & [1.000, 1.000]\\
rupture & 61 & 33/35 & [0.890, 0.996]\\
stroke \footnote[1] & 109 & 19/32 & [0.444, 0.743]\\
thrombosis & 51 & 30/31 & [0.927, 1.000]\\
thrombus & 181 & 33/34 & [0.917, 1.000]\\
tumor & 127 & 39/42 & [0.862, 0.995]\\
\hline
\end{tabular}
\end{center}
\end{table}
\begin{table}[ht]
\caption{Samples and confidence intervals generated per abnormality after false positive reduction for Site 2}
\label{Table 5}
\begin{center}
\renewcommand{\arraystretch}{1.3}
\begin{tabular}{ | c| c | c | c |}
\hline
Keyword & \# Pos. Reports & precision & C.I. ($ \alpha = 95\% $)\\
\hline
acute ischemic event \footnote[1] & 5 & 0/5 & 0.000\\
aneurysm & 185 & 30/30 & [1.000, 1.000]\\
arachnoid cyst \footnote[2] & 0 & N/A & N/A\\
atrophy & 11421 & 30/30 & [1.000, 1.000]\\
bleeding & 32 & 29/32 & 0.906\\
calcification & 2800 & 30/30& [1.000, 1.000]\\
cancer & 38 & 29/30& [0.936, 0.998]\\
chronic ischemic event & 0 & N/A & N/A\\
cva \footnote[1] & 17 & 8/17& 0.471\\
cystic necrosis & 2 & 2/2 & 1.000\\
dissection & 7 & 6/7& 0.857\\
encephalomalacia & 889 & 30/30 & [1.000, 1.000]\\
fracture & 513 & 30/31& [0.905, 1.000]\\
glioblastoma & 21 & 21/21& 1.000\\
haemorrhage \footnote[2] & 0 & N/A & N/A\\
hematoma & 2197 & 39/40& [0.925, 1.000]\\
hemorrhage & 1396 & 48/51& [0.875, 1.000]\\
hernia & 89 & 32/32& [1.000, 1.000]\\
hydrocephalus & 377 & 30/30 & [1.000, 1.000]\\
hygroma & 167 & 31/31& [1.000, 1.000]\\
infarct & 6531 & 47/48& [0.937, 1.000]\\
ischemia & 1891 & 30/31& [0.904, 1.000]\\
ischemic change & 4678 & 30/30 & [1.000, 1.000]\\
malformation & 34 & 33/34& 0.971\\
mass & 2648 & 44/47 & [0.864, 1.000]\\
meningioma & 830 & 30/30& [1.000, 1.000]\\
polyp & 87 & 30/30& [1.000, 1.000]\\
polyposis & 3 & 3/3 & 1.000\\
rupture & 24 & 24/24 & 1.000\\
stroke \footnote[1] & 122 & 24/30 & [0.670, 0.930]\\
thrombosis & 32 & 32/32 & 1.000\\
thrombus & 92 & 30/31& [0.915, 1.000]\\
tumor & 114 & 33/34& [0.921, 1.000]\\
\hline
\end{tabular}
\end{center}
\end{table}
\footnotetext[1]{Abnormalities with confidence interval lower bound less than 85\%}
\footnotetext[2]{There were no reports positive for this abnormality in the dataset}
\end{document}